\documentclass[11pt]{article}
\usepackage[margin=1in]{geometry}
\usepackage{amsmath, amssymb}
\usepackage{graphicx}
\usepackage{setspace}
\usepackage{hyperref}
\usepackage[noabbrev]{cleveref}
\usepackage[numbers]{natbib}

\hypersetup{
    colorlinks=true,
    linkcolor=blue,
    citecolor=blue,
    urlcolor=blue
}

\title{A Spiking Sequence Generator for Polar Trajectories on Neuromorphic Hardware}

\author{William R.P. Nourse$^{1,2,*}$\\
\small ORCID: \href{https://orcid.org/0000-0002-1437-026X}{0000-0002-1437-026X}
\and
Roger D. Quinn$^1$\\
\small ORCID: \href{https://orcid.org/0000-0002-8504-7160}{0000-0002-8504-7160}
}

\date{
\small
$^1$Mechanical \& Aerospace Engineering, Case Western Reserve University, Cleveland, OH, USA\\
\vspace{0.5em}
$^2$Mechanical \& Aerospace Engineering, West Virginia University, Morgantown, WV, USA\\
\vspace{0.5em}
$^*$Corresponding author: \href{mailto:william.nourse@mail.wvu.edu}{william.nourse@mail.wvu.edu}
\vspace{0.5em}
\today
}

\begin{document}

\maketitle

\begin{abstract}
Neuromorphic controllers for size, weight, and power-constrained systems require neural architectures that are both energy-efficient and interpretable at the level of system dynamics. However, existing approaches either rely on end-to-end trained spiking networks with limited interpretability, or on converted classical controllers that fail to fully exploit neuromorphic dynamics. We present a spiking neural network (SNN) architecture for generating polar trajectories, using a winner-take-all (WTA) architecture with accessory populations that induce controlled transitions in neural activity. We demonstrate tuning rules for these population dynamics, and utilize a form of shunting inhibition to enable independent control of direction, speed, and radius of the resulting polar trajectories. We implement the network on the SpiNNaker2 neuromorphic processor, and demonstrate a two to three orders of magnitude reduction in wall-clock step time and three to four orders of magnitude reduction in energy expenditure when compared to conventional computing platforms.
\end{abstract}

\vspace{0.5em}
\noindent\textbf{Keywords:} winner-take-all; manifold; ring oscillator; neuromorphic engineering; event-based system; SpiNNaker2

\section{Introduction}

For running neural controllers on size, weight, and power (SWaP) constrained robotic platforms, interest is growing in using brain-inspired neuromorphic computing hardware as a promising alternative to graphics processing units (GPUs), due to their low latency and high energy efficiency~\cite{sandamirskya_2022}. Work has been done to develop controllers for neuromorphic hardware using spiking neural networks (SNNs), with most work being done at one of two extremes: training deep artificial neural networks in an end-to-end manner~\cite{stewart_2025,romero_2025,valles_2024}, or converting more classical control systems into SNNs which reproduce the desired dynamics~\cite{dewolf_2023,arana_2024}. While these approaches are individually valid and sound, they come with tradeoffs for robotic applications. For end-to-end networks trained with gradient-based optimizers, questions of interpretability, safety, and stability arise as the networks usually behave as a black-box with little capability of human understanding. On the other extreme, converted classical controllers typically encounter approximation errors from discrete spike-timing, and provide minimal value when compared to modern low-power processors, which can evaluate classical control algorithms swiftly and efficiently. Ideally, networks deployed on neuromorphic hardware would offer a compromise somewhere in-between these two extremes. Neuromorphic hardware is capable of generating rich temporal dynamics beyond traditional feedforward neural networks, and networks should be capable of utilizing these dynamics while allowing controllable interactions. These systems should also be able to use latent representations which are human interpretable, and compatible with feedback control. In total, the field needs a network solution which preserves low-dimensional interpretability, enables explicit control of trajectory dynamics, and maps efficiently to neuromorphic hardware.

Within neuroscience, one approach which fulfills these criteria is that of \textit{neural manifolds}~\cite{perich_2025}. The key insight of this area of work is that the primary variation in activity of large populations of neurons can often be reduced to a small number of degrees of freedom, using dimensionality reduction techniques such as principal component analysis. For neural circuits involved with behavior, these can often be expressed as trajectories in as few as two dimensions, which can exhibit state-dependent dynamics including fixed points and limit cycles. These low-dimensional trajectories can be interpreted as dynamical primitives~\cite{ijspeert_2013}, enabling the translation of large-scale neural activity into combinations and variations of desired behaviors.

While manifolds can have many dimensions, many manifolds across the animal kingdom often reduce to a polar or near polar representation. Examples of these include orientation and head-direction estimation~\cite{kim_2016}, pattern generation for legged locomotion~\cite{linden_2022}, and the control of peristalsis~\cite{riddle_2022} and reaching~\cite{churchland_2012}. While these manifold trajectories resemble simpler systems such as the Hopf oscillator~\cite{khalil_2002}, their dynamical behavior is more rich due to how the trajectories can be modulated both spatially and temporally by descending influences and sensory feedback. As each of these behaviors is related to a common and useful task for robotic systems, this motivates the development of neural architectures to generate these behaviors which mimic the latent manifold structure observed in \textit{in-vivo} neural recordings. Rather than modeling a specific biological circuit, we abstract these observations into a design principle: low-dimensional, controllable neural dynamics that can be directly implemented on neuromorphic hardware.

Given that observable manifolds act as a low-dimensional dynamical system, an argument could be made that they could be reasonably implemented via training of randomly-connected recurrent neural networks such as echo-state networks or liquid state-machines~\cite{maass_2011,mantas_2012}. While recurrent reservoirs are known to be universal approximators of dynamical systems~\cite{funahashi_1993}, that does not imply that the resulting implementation within the reservoir is space-efficient. Indeed, the balance between memory capacity and ability for nonlinear processing is constrained by the size of the reservoir~\cite{dambre_2012,jaeger_2001}, meaning that sufficient representations may require prohibitively large reservoirs. This constraint is particularly problematic for implementation on neuromorphic hardware, as neuromorphic systems are more constrained on compatible network sizes due to their co-located computation and memory. For neuromorphic systems then, a structure which is constrained by functionality is desired.  

One family of neural architectures for generating low-dimensional manifolds of neural activity are variants of attractor networks, where constraints on neural connectivity and parameters cause the population-level dynamics to converge to well-defined behavioral regimes~\cite{khona_2022}. To replicate the behavior observed in biological manifolds, the network must be conducive to stable activity within one region of neural state space, and be transformable into sequential activity via controlled perturbations. Many techniques and topologies have been described to generate attractor networks with these properties, including the winner-take-all (WTA)~\cite{maass_2000}, ring attractor~\cite{seung_1998}, dynamic field theory~\cite{sandamirskya_2010,sandamirskya_2014}, winnerless competition and stable heteroclinic channels (SHCs)~\cite{rabinovich_2006,horchler_2015}, chains of Hodgkin-Huxley neurons~\cite{huo_2026}, and combinatorial threshold-linear networks~\cite{parmelee_2022}, among others~\cite{lehr_2024}. Each of these specific architectures causes sequential behavior through different dynamical regimes, ranging from sequential switching in WTA and SHC networks to a more continuous representation in ring attractors. The head direction circuit located within the \textit{Drosophila} central complex in particular has been simulated using spiking neurons organized into a ring attractor~\cite{stentiford_2024}, demonstrating the validity of this approach. The WTA network on the other hand, while more discrete than the ring attractor, is simpler to implement due to the robustness and simplicity of its connectivity patterns. All of these networks operate under different fundamental regimes, but the ultimate result is a qualitatively similar rotating sequence of activity among the neurons in the network.

Compared to traditional computing hardware solutions such as CPUs and GPUs, neuromorphic hardware offers a low-energy solution which is well-suited to simulating attractor networks under SWaP constraints, including the WTA. Neuromorphic architectures are naturally designed for implementing networks of spiking neurons which only communicate with spatiotemporally sparse patterns of binary events. This enables higher energy efficiency for simulating WTA dynamics over other devices which require all elements of the network to be computed at all times. Accordingly, the WTA architecture has been implemented in neuromorphic hardware numerous times, including FPGA systems~\cite{abdoli_2020, you_2021}, VLSI implementations~\cite{you_2017,camilleri_2010,giulioni_2012,locatelli_2025,giulioni_2015}, mixed-signal architectures~\cite{barranco_2022}, memristor and spintronic devices~\cite{wang_2019,wang_2020,wang_2023}, as well as the first generation SpiNNaker processor~\cite{denk_2013,osorio_2024,schoepe_2019}. However, most of these applications have focused on the use of WTA networks for sensory integration and decision-making, as opposed to controllable sequences of activity.

In this work, we develop a spiking neural network (SNN) architecture for generating controllable sequences of neural activity based on the WTA framework. We choose a WTA-based formulation over a continuous ring attractor, trading angular resolution for controllability and hardware efficiency. A recurrent population of leaky-integrate and fire (LIF) neurons with WTA dynamics receives controlled perturbations from accessory populations to induce polar rotations within the population, and a readout population enables independent control of the trajectory output radius. We implement this system onboard the SpiNNaker2 neuromorphic processor~\cite{hoppner_2021}, due to its event-driven energy efficiency and its optimized design for large-scale networks of dynamical neurons with arbitrary connectivity. Recent work has demonstrated the validity of using perturbed WTA dynamics in spiking neurons to form a ring oscillator~\cite{huo_2026,huo_2026b}. However, our approach differs from theirs in terms of the neural dynamics used, the addition of external populations for controlling the output trajectory, as well as our implementation on physical neuromorphic hardware. We also perform benchmarking analyses of our neuromorphic solution, demonstrating that the SpiNNaker2 neuromorphic system enables inference of WTA-style networks with orders of magnitude lower energy usage and simulation time than traditional computational hardware.

\section{Methods}\label{sec:neuron}
\subsection{Neural Dynamics}
For all neurons in the network, we use a discretized leaky-integrate and fire (LIF) model,
\begin{equation}\label{eq:membrane}
    V[t+1] = \beta\cdot V[t] + S[t+1] + B,
\end{equation}
where $V$ is the membrane potential, $\beta\in [0,1)$ is the membrane decay factor, and $B$ is a constant bias. We denote the firing of a spike as $\delta(t)$,
\begin{equation}\label{eq:threshold}
    \delta(t) = 
\begin{cases}
    1, & V(t)\geq\theta\\
    0, & \text{otherwise},
\end{cases}
\end{equation}
with $\theta$ being the voltage threshold for generating a spike. Upon generating a spike, the membrane potential is reset to 0. $S$ is the sum of all weighted input spikes after filtering through exponential synapses, with two dynamical variables corresponding to excitatory ($S_A$) and inhibitory ($S_B$) synapses.
\begin{equation}\label{eq:excitatory exponential synapse}
    {S}_{A}^i[t+1] = \alpha_{A}\cdot S_{A}^i[t] + \sum_j^N w_{A}^{ij}\cdot\delta_j[t],
\end{equation}
\begin{equation}\label{eq:inhibitory exponential synapse}
    {S}_{B}^i[t+1] = \alpha_{B}\cdot S_{B}^i[t] + \sum_j^N w_{B}^{ij}\cdot\delta_j[t],
\end{equation}
\begin{equation}\label{eq:exponential synapse}
    S[t] = S_{A}[t] - S_{B}[t]
\end{equation}
where $w_{ij}$ is the weight from presynaptic neuron $j$ to postsynaptic neuron $i$, $\alpha\in [0,1)$ is a synaptic decay factor, and $\delta_j[t]$ is the spiking output of neuron $j$. 

Neurons also have a counter tracking the status of their refractory period. Once a neuron spikes, it enters the refractory period and the counter is reset to its maximum length $T_{refrac}$. For every step where the counter is nonzero, the counter is decremented by one and the membrane update is replaced with $V[t+1]=V[t]$, holding the membrane at rest until the refractory period is over. Synaptic dynamics are allowed to continue during the refractory period.

\begin{figure}[t]
 \centering
        \includegraphics[width=0.9\textwidth]{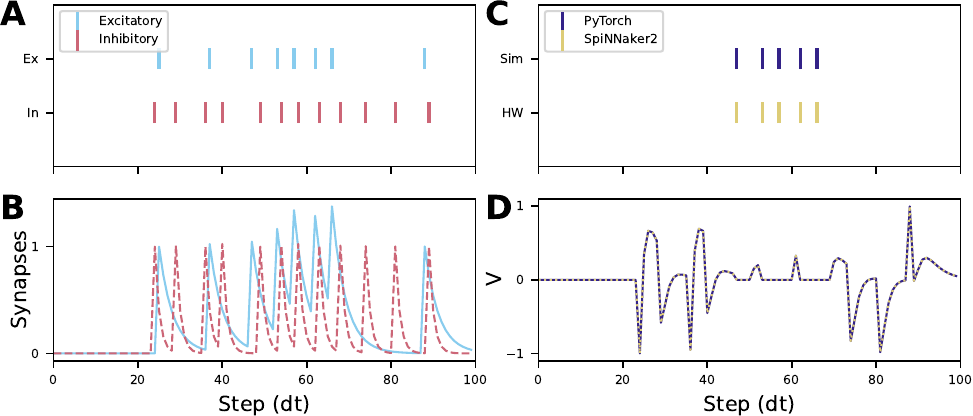}
 \caption{Neural model used in this work. (A) Excitatory (top) and inhibitory (bottom) input spikes to the model, and (B) the corresponding filtered spike trains. (C) Spikes generated by the simulation model (top) and the hardware (bottom). (D) Voltage trace of the simulation and hardware (dashed) models.}
\label{fig:neuron}
\end{figure}

We implemented~\cref{eq:membrane,eq:threshold,eq:excitatory exponential synapse,eq:inhibitory exponential synapse,eq:exponential synapse,eq:gain synapse} in the machine learning framework PyTorch~\cite{paszke_2019}, before deploying to the SpiNNaker2 neuromorphic processor. \textbf{Figure~\ref{fig:neuron}} compares the spike times and voltage traces of the simulated and hardware implementations, showing zero discrepancies from step to step. This identical behavior allows us to perform initial development in simulation before deploying to the hardware.

\subsection{Modulatory Synaptic Dynamics}
As currently described, the synapse model described in \cref{eq:excitatory exponential synapse,eq:inhibitory exponential synapse,eq:exponential synapse} is purely additive. For our proposed network architecture, we desire the ability to modify the effective gain of some neurons over the course of inference. Inspired by the multiplicative effects of shunting inhibition~\cite{groschner_2022,nourse_2023}, we introduce a gain modulating synaptic variant, $S_{gain}$, where two channels $S_A$ and $S_B$ are multiplied instead of being subtracted:
\begin{equation}\label{eq:gain synapse}
    S_{gain}[t] = S_{A}[t] \cdot S_{B}[t]
\end{equation}
This allows synaptic influence to modify the effective gain of the neuron, similar to the gain adjustment mechanism in the spiking elementary motion detector~\cite{milde_2018}. We implemented these dynamics in a custom neural model in the py-spinnaker2 API, enabling execution on the SpiNNaker2 hardware.

\begin{figure}[t]
 \centering
        \includegraphics[width=0.9\textwidth]{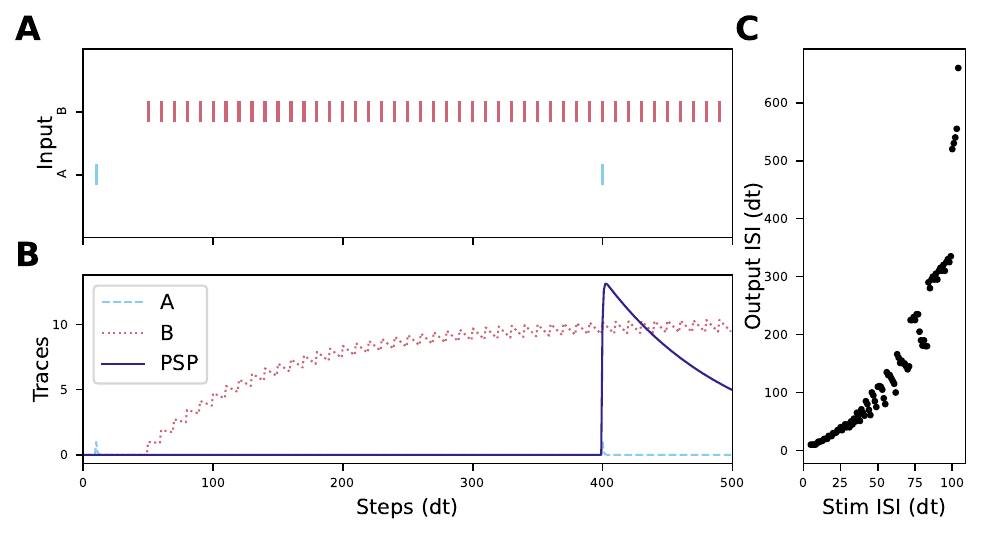}
 \caption{Multiplicative exponential synapses. Incoming spikes (A) are routed into either channel based on their associated weight. The exponential synaptic traces are multiplied together (B), such that overlapping activity in both inputs is required to elicit an excitatory post-synaptic potential (PSP). (C) The exact inter-spike interval (ISI) of the neuron given steady-state firing from its inputs can be found empirically as a function of the incoming modulatory ISI, and follows a roughly exponential relationship. All data shown were recorded in simulation using PyTorch.}
\label{fig:shunt}
\end{figure}

\textbf{Figure~\ref{fig:shunt}} illustrates the behavior of this multiplicative synapse. As the inter-spike interval of the modulatory input changes, the output firing interval varies following an approximately exponential relationship. This demonstrates that the synapse provides a controllable gain mechanism, enabling continuous modulation of neuronal excitability through spike timing alone. 

\subsection{Hardware}\label{sec:hardware}
To deploy the spiking sequence generator network for inference, we used the SpiNNaker2 neuromorphic processor~\cite{hoppner_2021} on a single-chip development board (SpiNNcloud Systems GmbH, Dresden, Germany). This is a massively-parallel embedded system, with 152 low-power ARM-based processing elements (PEs) connected within an asynchronous, event-driven mesh framework. We were able to use an existing model within the py-spinnaker2 API for the LIF model with exponential synapses, and implemented a custom variant to simulate the gain modulating synapse. Current support for synaptic weights on SpiNNaker2 with our chosen neural models is limited to signed 8-bit integers, so all analysis and tuning was done with awareness of this constraint. As each PE has one shared region of SRAM for model parameters, synaptic properties, and recording, we record all spikes and voltage traces to the provided 2 GB of external DRAM in order to maximize the available space for neural and synaptic information. Unless otherwise specified, all neural data presented in this work were recorded from execution on-board the SpiNNaker2 processor.

One challenging aspect of running large spiking sequence generators on SpiNNaker2 (or potentially any other digital neuromorphic system) is the constraint on synaptic fan-in and fan-out to any neural population. Unlike other systems such as Intel's Loihi family of processors~\cite{davies_2018}, SpiNNaker2 does not have hard limits on fan-in or fan-out tables. Rather, each processing element (PE) has a shared memory region (approx. 92kb) in which to store the neural parameters, neural and synaptic states, and synaptic connectivity information. This means that each PE can either simulate a large number of neurons with sparse synaptic connectivity, or a small number of neurons with dense connectivity. As part of our model requires self-recurrent inhibition within a winner-take-all (WTA) population, to maximize the potential synaptic connectivity we distribute the WTA neurons evenly across as many PEs as possible. We set the maximum number (or atomic maximum) of WTA neurons per PE $A_{WTA}$ as
\begin{equation}\label{eq:map}
    A_{WTA} = \lceil \frac{N}{P-\lceil\frac{N}{A_S}\rceil-\lceil\frac{N}{A_L}\rceil} \rceil < A_L,
\end{equation}
where $N$ is the number of WTA neurons, $P$ is the total number of user-available PEs (148 for SpiNNaker2 using the py-spinnaker2 API), $A_S$ is the atomic maximum of spike sources per PE, and $A_L$ is upper bound on the atomic maximum of LIF neurons with exponential synapses. This mapping ensures that the WTA neurons are distributed evenly across the PEs without exceeding per-PE memory limits.

\begin{figure}[t]
 \centering
        \includegraphics[width=0.9\textwidth]{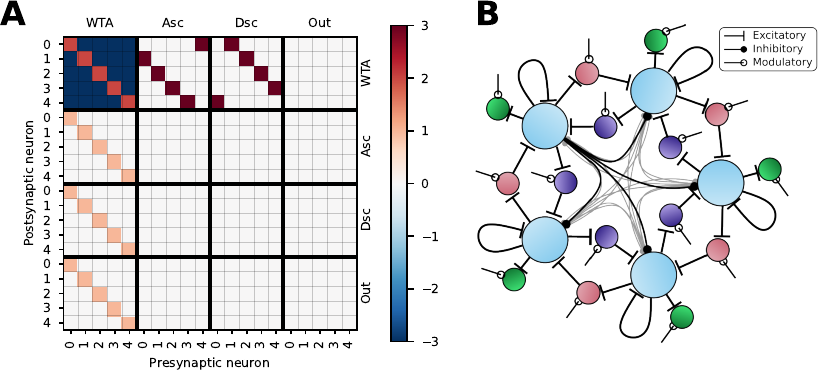}
 \caption{Overview of the entire neural architecture, in the form of a connectivity matrix (A) and schematic (B) for a sequence generator network with N=5 neurons in the winner-take-all (WTA) population.}
\label{fig:architecture}
\end{figure}

\section{Network Architecture and Tuning}
We present a spiking neural network (SNN) architecture for generating polar trajectories on neuromorphic hardware. This network consists of a spiking population with self-recurrence to form a winner-take-all (WTA) system, with accessory populations to facilitate controlled rotation and a readout population for output. \textbf{Figure~\ref{fig:architecture}} provides an overview of the full network architecture, with the connectivity matrix and schematic emphasizing how the WTA, accessory, and readout populations interact. This structure separates the stability (WTA), phase advancement (accessory populations), and amplitude control (readout), forming the basis for independently controllable polar trajectories.  We now present the design and validation of each of these components in the following subsections.

\subsection{Winner-Take-All Dynamics}
To construct a stable WTA system where one neuron is persistently active until another is sufficiently stimulated, the network needs two factors: self-excitation to maintain a desired baseline of activity, and global inhibition to suppress activity in all other neurons.

\begin{figure}[t]
 \centering
        \includegraphics[width=0.9\textwidth]{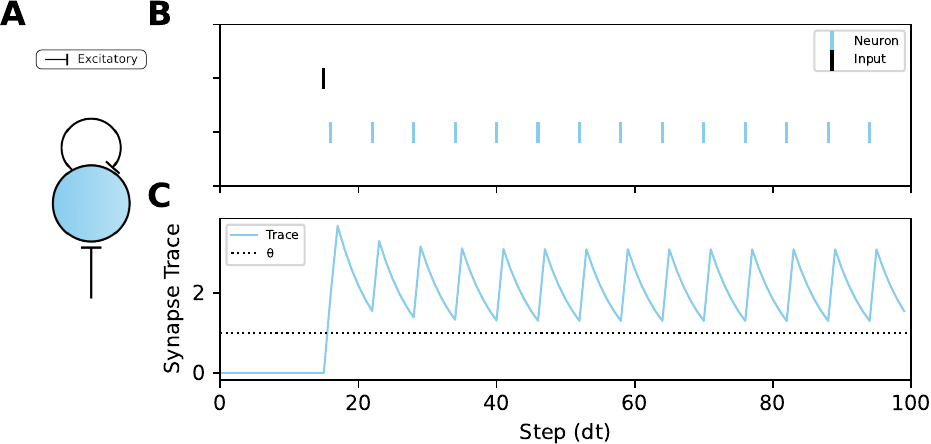}
 \caption{Neuron with stable self-excitation. (A) A single LIF neuron with exponential synapses receives external stimulation, and has a self-recurrent excitatory connection. When a single spike is received (B, top), the resulting change in the synaptic trace (C) is enough to induce the neuron to fire a spike. While the neuron is in its refractory period, the neuron is unable to fire a spike and the synaptic trace decays. Once the refractory period is over, the synaptic trace is still large enough to induce an immediate jump in membrane potential above the threshold, causing a new spike. This pattern continues in perpetuity, resulting in steady-state spiking (B, bottom). Spikes were recorded on SpiNNaker2, the synaptic trace was simulated as SpiNNaker2 does not support recordings of synaptic traces.}
\label{fig:self}
\end{figure}

\subsubsection{Self-Excitation}
We design recurrent excitation between each neuron and itself, such that the neuron fires at a desired rate once it receives sufficient external stimulation. Our goal is to have a neuron, once stimulated, firing at a constant inter-spike interval $T_{ISI}$. We use a standard LIF model with exponential synapses (\cref{eq:membrane,eq:threshold,eq:excitatory exponential synapse,eq:inhibitory exponential synapse,eq:exponential synapse}). To prevent excessive firing, we set the neuron's refractory period as $T_{ref}=T_{ISI}-1$. Next, we ensure that the recurrent spike is self-sufficient to generate future activity by setting the recurrent weight $w_{ii}>\theta$. In practice, this means that $w_{ii} = \lceil\theta+\epsilon\rceil$, where $\epsilon\ll1$ is a small constant and $\lceil \rceil$ is the ceiling function.

Finally, we need to tune the excitatory synaptic decay such that the neuron is able to fire once the refractory period has elapsed. After the initial spike, the synaptic activation decays with rate $\alpha_{ex}$ such that
\begin{equation}
    S_{ex}[t] = w_{ii}\cdot\alpha_{ex}^{t-1}.
\end{equation}
After the refractory period, we need to ensure that the synaptic activation is still high enough to induce a spike. We frame the inequality
\begin{equation}
    w_{ii}\cdot\alpha_{ex}^{T_{ISI}-1} >\theta,
\end{equation}
and solve for $\alpha_{ex}$ to find that
\begin{equation}
    \alpha_{ex}>\left ( \frac{\theta}{w_{ii}} \right ) ^ {1/(T_{ISI}-1)}
\end{equation}

\textbf{Figure~\ref{fig:self}} presents a neuron with self-excitation tuned following these equations, which undergoes sustained periodic spiking after a single input of sufficient magnitude. This persistent activity demonstrates that the recurrent excitation alone can maintain stable firing, and consequently can maintain an active state within our WTA network.

\begin{figure}[t]
 \centering
        \includegraphics[width=0.9\textwidth]{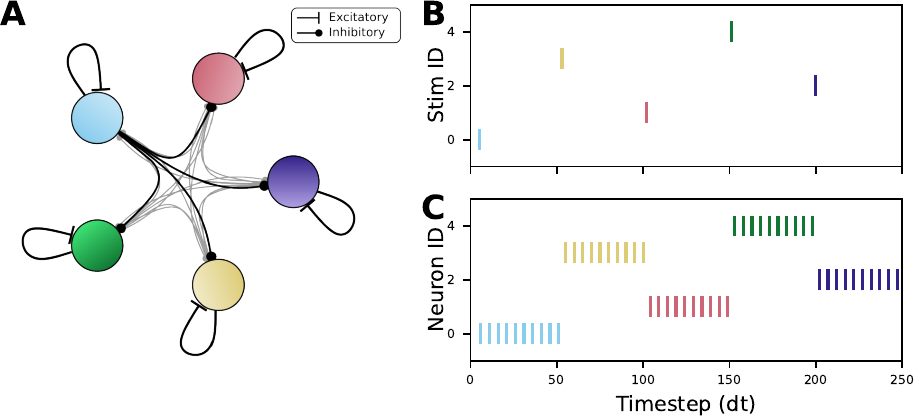}
 \caption{Winner-take-all (WTA) architecture, pictured with N=5 neurons. (A) Neurons with self-excitatory recurrence are arranged topologically in a ring, with inhibitory connections to every other neuron in the ring. Inhibition is tuned such that once a neuron receives an external stimulus (B), then that neuron starts to fire and suppresses activity in all other neurons (C). In this manner, there is only one active neuron in the network at any time. All activity was recorded on SpiNNaker2.}
\label{fig:wta}
\end{figure}

\subsubsection{Global Inhibition}
Once a single neuron is spiking, global inhibition is needed to ensure that no other neurons are able to spike. To tune this global inhibition $w_{i\neq j}$, we examine the case where a new neuron has been stimulated and needs to fire, suppressing the currently active neuron. This requires the magnitude of the inhibitory synaptic trace to be greater than that of the excitatory trace. During steady-state firing, the value of the excitatory trace at the moment of firing is
\begin{equation}
    S_{exp}[T_{ISI}] = \left ( S[T_{ISI}]+w_{ii} \right ) \cdot\alpha_{ex}^{T_{ISI}},
\end{equation}
therefore we find that the condition for successful global inhibition is that
\begin{equation}
    w_{i\neq j}>\frac{w_{ii}\cdot\alpha_{ex}^{T_{ISI}}}{1-\alpha_{ex}^{T_{ISI}}}+w_{ii}
\end{equation}
In \textbf{Figure~\ref{fig:wta}}, we demonstrate the WTA dynamics within the network. Once any neuron becomes active, it suppresses all other neurons via global inhibition. This ensures that only one neuron is active at a time, creating a simplified state representation which we can use for mapping network activity to angular positions in polar space.

\subsection{Rotation and Speed Control}\label{sec:rotation}
Once we have stable attractor dynamics in the WTA system, the next step for generating usable polar trajectories is to allow controlled rotation in the neural state-space. Inspired by the protocerebral bridge in the insect central complex~\cite{kim_2016}, we add two accessory rotation populations to our WTA network (\textbf{Figure~\ref{fig:rotation}A}). These populations are the same size as the WTA and enable selective gating of neural activity within the network, allowing an autonomous rotation from one active node to the next in a desired direction after that node has been active for a prescribed duration (\textbf{Figure~\ref{fig:rotation}C-H}). Translating the population activity into polar coordinates, we allow each neuron in the WTA to indicate a discrete angle in polar space(\textbf{Figure~\ref{fig:rotation}B}).

\subsubsection{Rotational Connectivity}
By itself, the WTA network maintains a stable region of activity which does not drift or decay due to the self-excitation and symmetric global inhibition. To transform a WTA network into a system which generates a sequence of activity, the connectivity pattern needs to be shifted asymmetrically such that there is an imbalance between excitation and inhibition. By biasing this imbalance towards the desired direction of travel, this asymmetry causes neural activity to propagate throughout the network, while retaining the core WTA behavior of only one neuron being active at a time. When the population activity is viewed in polar coordinates, this manifests as rotational dynamics.

\begin{figure}[!tbp]
 \centering
        \includegraphics[width=0.9\textwidth]{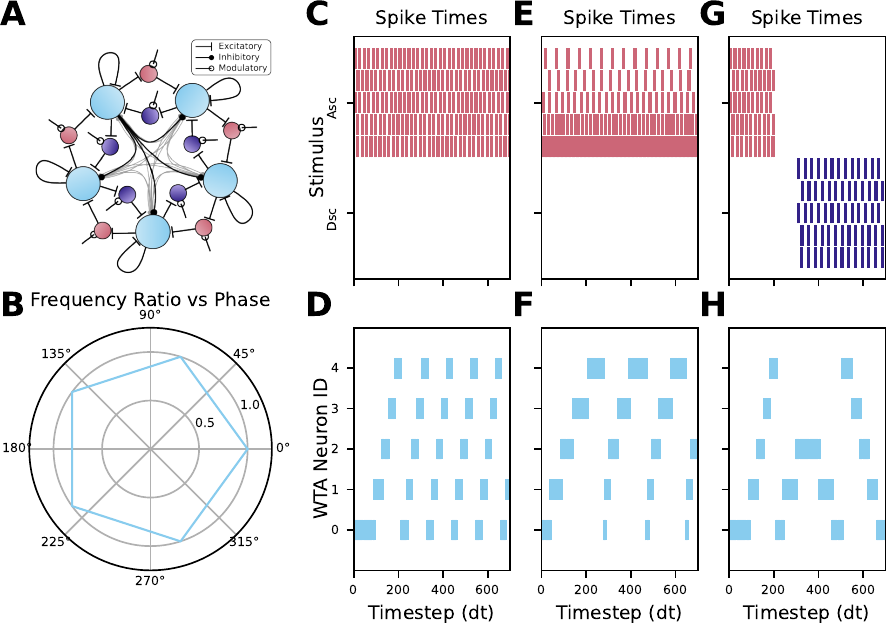}
 \caption{Accessory populations facilitate controlled perturbations for system-wide rotation in a winner-take-all (WTA) network with $N=5$ WTA neurons. (A) Schematic representing the rotational mechanism.(B) We can plot the activity of the rotating WTA network in polar coordinates. The identity of the active neuron corresponds to a specific angular range $\phi$, and the radius $r$ is the firing frequency scaled by the maximum firing rate. Stimulation applied to the ascending rotation population (C, E, G, top/red) induces rotation with neurons of increasing index, and stimulation to the descending population (C, E, G, bottom/purple) performs the opposite. Rotation can be seen in the WTA population (D, F, H) as a sequence of activity. Constant input to the ascending population (C, D) causes rotation with uniform speed. Varying the stimulation level to each individual rotation neuron (E, F) causes a likewise shift in activation period across the WTA, stabilizing after an initial warmup period. Changing stimulation from the ascending to descending populations (G,H) causes the network to stop rotating, then switch directions. All spiking activity was recorded on SpiNNaker2.}
\label{fig:rotation}
\end{figure}

To control these rotational dynamics during operation, instead of changing the true connectivity matrix we instead change the effective connectivity through the introduction of two accessory populations. Each of these populations is of size $N$, and exchanges reciprocal excitation with the WTA network (\textbf{Figure~\ref{fig:rotation}A}). When the currently active neuron in the WTA fires, it excites the corresponding neurons in both accessory populations, until one accessory neuron is sufficiently activated to spike. This spike is transmitted to the neighboring neuron in the WTA, perturbing the system such that a new neuron is active and the original neuron is silenced. Both accessory populations have the same neural parameters, and the only differences in connectivity are which neurons in the WTA are excited (the ascending population excites the neuron of index $i+1$, and the descending population excites $i-1$). The synaptic weight from the accessory populations to the WTA network is chosen such that it overcomes the inhibition experienced by the target neuron.

\subsubsection{Modulatory Gating}
Without external control, both accessory populations would become active simultaneously and destroy any activity in the WTA network. For rotation to be possible and controllable, we require a mechanism for gain control of excitability in the accessory populations. To do this, each neuron in the rotation populations is a LIF model with multiplicative exponential synapses (\cref{eq:gain synapse}). These neurons receive two synaptic inputs, one excitatory input from their corresponding neuron in the stable attractor, and another modulatory external input which controls the rotation speed. We select a high decay factor for the gain synapse $\alpha_{gain}=0.99$ so that the synaptic gain trace approximates a steady-state value, which changes in magnitude based on the rate of external stimulation. For the rotation neuron itself, we also set a high neural decay $\beta_{rot}=0.99$ to ensure that incoming synaptic currents are integrated without significant losses between steps. We then set the firing threshold $\theta_{rot}$ empirically, by finding the steady-state voltage the neuron reaches with unity gain and repetitive excitation from the stable attractor with interval $T_{ISI}$, and subtracting a small factor $\epsilon\ll 1$ to ensure spiking. 

Controlling the ISI of the modulatory spike train into the rotation neuron dictates the gain of the synaptic dynamics, and consequently the speed at which the overall system transitions from one active neuron to the next. Through numerical simulations, we can see that this relationship follows an approximately exponential trend (\textbf{Figure~\ref{fig:shunt}C}), and we can use this tuning curve to set the firing rate accordingly for a desired population rotation speed (\textbf{Figure~\ref{fig:rotation}C-D}). This relationship is also spatially decoupled, meaning that changing the modulation interval for each individual rotation neuron allows arbitrary control of the rotational speed at every trajectory angle (\textbf{Figure~\ref{fig:rotation}E-F}). Similarly, changing which rotation population is stimulated dictates the direction of rotation (\textbf{Figure~\ref{fig:rotation}G-H}).

\begin{figure}[t]
 \centering
        \includegraphics[width=0.9\textwidth]{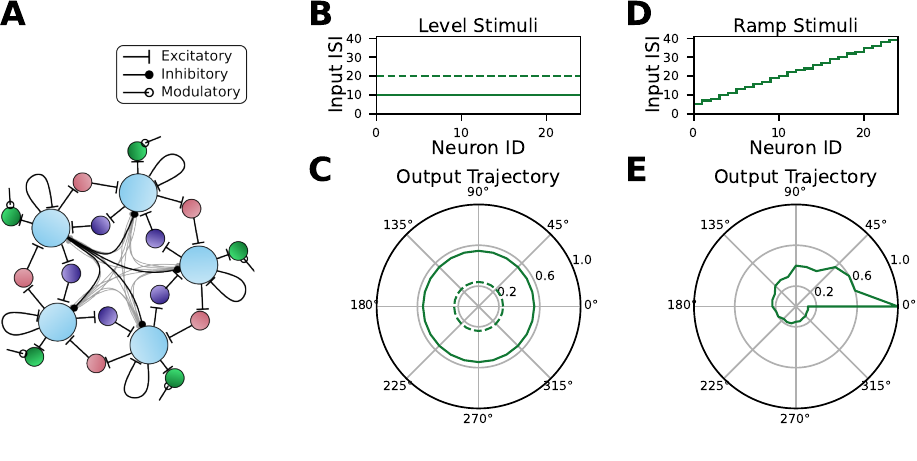}
 \caption{Modulation of a readout population allows control of polar radii. (A) Schematic of the readout architecture. Each neuron in the winner-take-all (WTA) circuit excites a single neuron in the readout population. Each neuron in the readout population also receives modulatory input from external stimulation. By plotting the output activity in polar coordinates, with the active neuron corresponding to the phase and the readout firing rate controlling the radius, we generate a polar trajectory. (B) Driving the readout population with a smaller ISI (higher frequency) causes a higher radius, and raising the ISI (lowering the frequency) causes the radius to decrease. (C) This stimulation is decoupled from neuron to neuron, allowing individual control of the resulting radii from every neuron in the WTA. All polar trajectories are derived based on recorded spiking activity on SpiNNaker2.}
\label{fig:shape}
\end{figure}

\subsection{Shape Control}\label{sec:shape}
Adding rotational accessory populations to the WTA enables the control of timing and rotational direction within the population. As the identity of the active neuron controls the trajectory angle, the firing rate of said neuron dictates the radius. However, directly modulating the firing rate of neurons within the WTA network would also affect the rotational timing of the network itself, which is antithetical to our goals. Instead,we decouple amplitude from phase by introducing a readout population which scales activity from the WTA by an external modulatory signal.

For our readout implementation, we take inspiration from the work of Ajallooeian et al.~\cite{ajallooeian_2013}, who demonstrated the design of oscillators with arbitrary limit cycle trajectories by starting with a base oscillator with a simple trajectory, and applying a mapping function with radial basis to scale the trajectory into an arbitrary polar form. We treat the rotational trajectory formed by our coupled WTA network and rotational populations as a rhythmic reference, and add a readout population of LIF neurons with multiplicative exponential synapses (\cref{eq:gain synapse}). Each readout neuron receives excitatory input from one neuron in the stable attractor, and modulatory input from external stimulation (\textbf{Figure~\ref{fig:shape}A}). By changing the rate of modulatory input, we have control of the firing rate of each neuron individually in the readout population. This is analogous to theories of spinal locomotion control, where spinal interneurons are divided into a global rhythm generation network and individual pattern formation circuits for the various muscle groups~\cite{rybak_2006}.

Similar to the rotational populations described in Section~\ref{sec:rotation}, the ISI of modulatory external stimulation to the readout neurons directly controls the output firing rate, and consequently the radii along the polar trajectory. This modulation is also decoupled from neuron to neuron, allowing fine-grained spatial control of the radius at every angle. We demonstrate this in \textbf{Figure~\ref{fig:shape}}, where we show the resulting trajectories from two different global stimulation levels, as well as a linear ramp distributed across the entire population. These examples confirm that by varying the level modulation of the readout layer, we can control the radii of the polar trajectory without affecting the rotational dynamics.

\begin{figure}[t]
 \centering
        \includegraphics[width=0.9\textwidth]{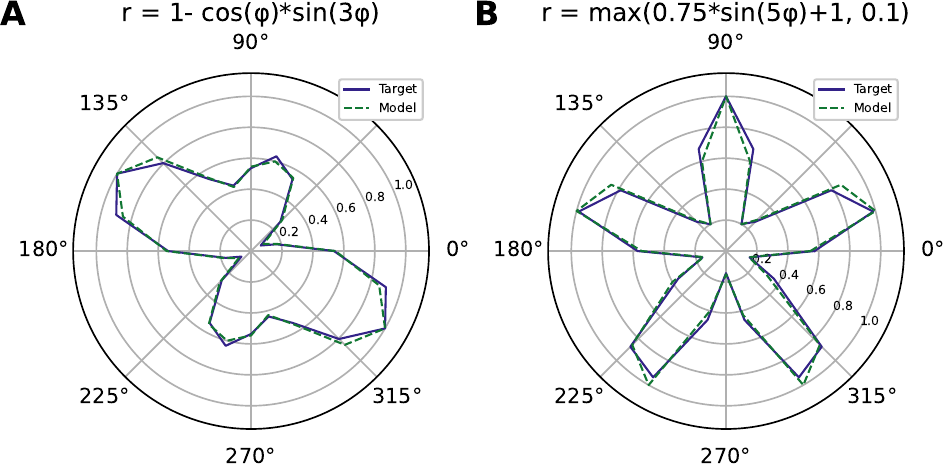}
 \caption{Controlling the external stimulation of the readout population allows arbitrary control of the trajectory radius. This is amenable to simple optimization, as the resulting firing rates are dependent on the incoming inter-spike intervals (ISIs), which are constrained to a small and finite space by the discrete simulation dynamics. The output firing rate of each neuron in the readout population with varied stimulation is evaluated in parallel, and the mean-squared error between the actual and desired rates can be evaluated quickly via a grid search. By selecting the stimulation rates which produce the least error per neuron, an optimized polar trajectory can be produced such as a polar curve with harmonic modulation (A) or a five-petal rose curve (B). Batch processing was performed in simulation using PyTorch.}
\label{fig:fit}
\end{figure}

\subsection{Optimizing Trajectories}
Given that it is possible to directly control the radius at every point, we can optimize the stimulation parameters to generate a desired trajectory in polar coordinate space. As the ISI space is integer-bound due to hardware temporal discretization, and the PyTorch version of the complete network fits comfortably within consumer GPU VRAM, we can perform the optimization via brute force in a time-efficient manner. Using our PyTorch simulation model (see Section~\ref{sec:neuron}), we simulate M parallel copies of our network with N WTA neurons, where M is the usable space of integer ISI values ($M\in[5,50)$), and we evaluate these M copies in a single batch on a consumer GPU. Reading out the resulting ISIs, we can directly apply a mean-squared error loss function and find the desired stimulation ISIs in a single step. This approach leverages the discretized temporal resolution of the hardware, converting trajectory synthesis into a finite combinatorial search problem. Because the mapping from input ISI to output firing rate is low-dimensional, this search scales linearly with the number of neurons and does not require iterative training. Examples of this process performed for two distinct polar functions can be seen in \textbf{Figure~\ref{fig:fit}}, showing that the network can be configured to produce complex, user-defined trajectories in a single-shot manner.

\section{Hardware Performance}\label{sec:performance}
\subsection{Maximum Network Size}
\begin{figure}[t]
 \centering
        \includegraphics[width=0.9\textwidth]{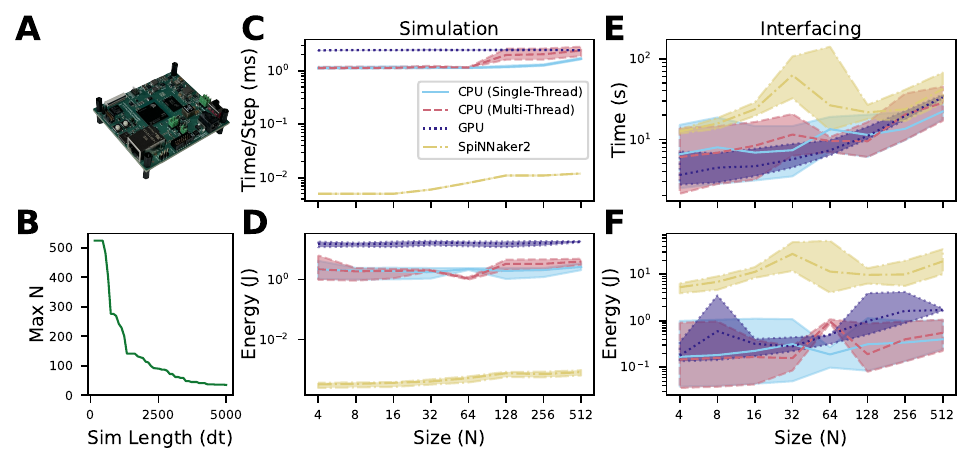}
 \caption{(A) Throughout this work, network simulations were performed on a single SpiNNaker2 Neuromorphic processor. (B) By performing a binary search using the SpiNNaker2 mapping algorithm, we collected the maximum network size (N WTA nodes) as a function of the total simulation steps. Simulating networks of increasing size, we recorded the mean wall-clock time per simulation step (C) and total energy for the simulation (D) for SpiNNaker2, as well as a consumer-grade CPU and GPU for comparison. We also recorded the total time (E) and energy (F) for loading data to and from these same platforms.}
\label{fig:hardware}
\end{figure}

 We determined the maximum network size which could run on the single SpiNNaker2 chip as a function of the number of simulation steps. In principle, the synaptic density could have an effect on the maximum allowable network size. However, for the single-chip system using our distributed mapping technique (Equation~\ref{eq:map}), we determined the main limiting factor to be the memory allocation for storing the input spike times for the four stimulus populations (one for the WTA, one each for the rotation populations, and one for the readout population). Note that this limitation arises from the current use of precomputed input spike trains, rather than intrinsic or closed-loop spike generation. To find the maximum network size for a given number of simulation steps, we exponentially increased the network size until the provided compiler failed to generate a successful hardware mapping. We then performed a binary search to find the largest size which successfully compiled. The results can be seen in \textbf{Figure~\ref{fig:hardware}B}, but in general we find that for small simulations (150 steps) we can simulate networks with 524 WTA nodes, with that decreasing to 35 nodes for longer simulations of 5000 steps. 

\subsection{Inference Performance}
To evaluate the hardware performance of our system when simulating our network, we performed a series of benchmarking experiments comparing our neuromorphic implementation to running the same simulation on a consumer-grade CPU or GPU. We simulated our network using four hardware configurations: a single-chip SpiNNaker2 system, an NVIDIA Gigabyte GeForce RTX 4070 GPU, an Intel Core i5-3570K Quad-Core CPU operating at 3.4 GHz, and the same CPU limited to using a single core for the simulation. For all configurations, we varied the network size by increasing powers of 2 from $N=4$ to $N=512$ and, after 50 initial warmup passes through the simulation, recorded the mean wall-clock time for simulating one step of the neural dynamics as well as the energy required for the total simulation (150 steps). We used the onboard energy recording functionality for SpiNNaker2, and used the Zeus energy profiling software~\cite{you_2023} for the CPU and GPU. For the SpiNNaker2 system, we also recorded the time and energy used in the interfacing overhead for the chip. This includes the wall-clock interfacing time, which includes the process of generating a network map, opening a transceiver between the host PC and the chip, loading the network onto the chip, and reading the data off of the chip at the end of the simulation, as well as the energy usage during these operations. As a comparable quantity for the CPU and GPU systems, we recorded the time and energy needed to generate and load the tensors onto said device. To compute the energy specifically used for inference on SpiNNaker2, we took the total energy used over the entire period of loading, running, and unloading the chip, and scaled it by the percent of time the chip spent in inference. We recorded from all five shunt resistors on the chip, monitoring the power consumption of the 0.5V, 0.8V, Phase-locked loop (PLL), I/O, and Q power lines.

To ensure we accurately captured the measurement variability, for each hardware configuration and network size we iteratively increased the number of trials $n$ until the half width of the data confidence interval was less than the data mean scaled by our target precision $\rho = 5\%$:
\begin{equation}
    \frac{1.96\cdot\sigma}{\sqrt{n}}<\rho\cdot\mu
\end{equation}
The number of required samples for each size and hardware configuration can be found in \textbf{Table~\ref{tab:samples}}. The CPU systems were more variable at smaller network sizes, while the GPU and SpiNNaker2 chip were fairly consistent across scales.

The results of our benchmarking experiments can be seen in \textbf{Figure~\ref{fig:hardware}C-F}. Across all network sizes, the SpiNNaker2 system performs each simulation step two-to-three orders of magnitude faster than a consumer CPU or GPU. The GPU simulated each network step at 2.4 milliseconds per step with little variation, and the single and multi-core CPU requiring between 1.1 milliseconds and 2.4 milliseconds per step. The SpiNNaker2 system ranged in simulation speed from five to twelve microseconds per step.

The SpiNNaker2 system used orders of magnitude less energy to complete the simulations than the other systems, using 0.3 to 0.8 millijoules. The GPU used the most energy of between 15 and 18 joules for the entire simulation. Both CPU systems used less energy than the GPU, ranging from 2.1 to 2.6 joules. 

\begin{table}[t]
\caption{Number of Samples Required for \textbf{Figure~\ref{fig:hardware}}}
\centering
\begin{tabular}{l r r r r r r r r}
\hline
Device & N=4 & N=8 & N=16 & N=32 & N=64 & N=128 & N=256 & N=512\\
\hline
CPU (Single-Thread) & 350 & 100 & 138 & 138 & 10 & 96 & 40 & 49 \\
CPU (Multi-Thread)  & 579 & 131 &  58 &  10 & 10 & 76 & 80 & 44 \\
GPU                 &  33 &  18 &  36 &  24 & 25 & 32 & 23 & 10 \\
SpiNNaker2          &  38 &  40 &  17 &  49 & 40 & 41 & 55 & 48 \\
\hline
\end{tabular}
\label{tab:samples}
\end{table}

While the SpiNNaker2 system demonstrated clear advantages for both inference time and energy, the benefits become less pronounced once the time and energy used for interfacing with the hardware are included. In this scenario, all of the hardware options have loading times on the same order of magnitude with SpiNNaker2 taking the longest. Consequently, the single-chip SpiNNaker2 system uses one to two orders of magnitude more energy than the CPU and GPU systems for interfacing with the hardware. While this initialization overhead dominates these short simulation runs, the neuromorphic advantages will dominate for sustained and embedded deployments.

The comparisons in this section reflect implementations of the same discrete-time spiking dynamics across platforms, rather than fully optimized hardware-specific kernels. While further optimization could reduce the absolute runtime and energy usage on traditional hardware, the relative energy and latency advantages of the neuromorphic hardware are expected to persist.

\section{Discussion and Future Work}
 This work establishes a design pattern for constructing controllable, low-dimensional dynamical systems on neuromorphic hardware, using a spiking neural architecture which generates sequences of neural activity. The resulting sequences can be mapped into a polar-coordinate space, allowing for the arbitrary and independent control of speed, direction, and radius of the resulting trajectories. These trajectories can directly parameterize rhythmic controllers for locomotion, scanning behaviors, or sensorimotor coordination in embedded robotic systems. We implemented the network on a single SpiNNaker2 neuromorphic processor, and compared the inference time and energy performance against a consumer-grade CPU and GPU. We found that the neuromorphic solution performs the simulation in orders of magnitude less time and energy, although initialization overhead is a significant performance barrier for short simulations.

As described in the Introduction, the inspiration for this work was the qualitative similarity between manifolds observed in the nervous system, across both phyla and behavior. While the dominant behavior of these natural manifolds can be explained in two-dimensions, many of these manifolds also vary in a third dimension. Expanding the connectivity of our network from a ring to a toroidal architecture would enable the generation of trajectories in spherical coordinates, moving beyond the polar trajectories we have shown.

The sequences in this work are generated by a winner-take-all (WTA) network, which constrains activity to a single active neuron at any time. This enables spatially decoupled control inputs, but limits the angular resolution of the resulting polar trajectories. Biological systems, such as the \textit{Drosophila melanogaster} central complex~\cite{kim_2016}, avoid this discretization by employing continuous ring attractors. While such attractors provide fine-grained angular representations, their overlapping connectivity introduces strong coupling between neighboring neurons, complicating independent modulation of trajectory parameters. In contrast, our WTA formulation enables per-state control with minimal coupling, at the cost of discretized angular resolution.

Our methodology is similar to the strong body of work in networks with winnerless competition~\cite{rabinovich_2006}, particularly coupled networks of stable heteroclinic channels (SHCs)~\cite{horchler_2015}. Future work could explore mapping relationships between the leaky integrate and fire dynamics commonly implemented on neuromorphic systems such as SpiNNaker2, and the Lotka-Volterra dynamics typically employed in SHC-based controllers. This mapping would allow the use of the analysis tools available to designing SHC systems to spiking neural networks (SNNs), and would likewise enable neuromorphic hardware to be used for more complex robotic control algorithms such as Dynamic Movement Primitives~\cite{rouse_2024}.

We modified an existing neural model implemented on SpiNNaker2 to have synaptic gain control. Due to the current high-level software infrastructure on SpiNNaker2, this ``gain" synapse replaces one of the existing synapses. This means that any neurons with external gain control can only receive excitation or inhibition, not both. Modifying the communication infrastructure within the high-level SpiNNaker2 API to support three synaptic channels would enable more complex control, allowing different accessory populations to inhibit one another and perform other gating functions.

During our performance testing in Section~\ref{sec:performance}, the main barrier we encountered in the maximum possible size of the network was the amount of space required for storing the stimulation spike times. We chose to use stored lists of spikes over spikes procedurally generated by Poisson inputs, as they offered fine-grained control for the experiments in Section~\ref{sec:rotation} (specifically changing the direction halfway through a simulation). With closed-loop spike generation or procedural input models, this memory bottleneck would be removed, enabling substantially larger and longer-running networks. Importantly, the core WTA and accessory dynamics themselves do not impose this scaling limitation. As of the writing of this manuscript, the API for interacting with SpiNNaker2 does not yet support evaluating neurons with exponential synapses in an iterative, closed-loop fashion. Once that functionality is complete, we anticipate that larger networks could be simulated for nearly infinite durations. Additionally, performing similar benchmarking experiments on a larger multi-chip SpiNNaker2 system could reveal the network scale at which the global inhibitory connectivity becomes the primary scaling issue.

While WTA networks and ring oscillator models have been studied for decades, this work provides a concrete methodology for implementing a controllable WTA-based architecture on the SpiNNaker2 neuromorphic processor. The network supports online, spike-based modulation of all trajectory parameters, enabling integration into larger spiking control systems. In contrast to trained recurrent networks or continuous attractor models, our approach emphasizes explicit controllability, parameter interpretability, and hardware efficiency. Although it introduces discretized angular representations and is currently constrained by implementation-level memory limits, it enables deterministic trajectory synthesis and direct deployment on neuromorphic hardware. Future work will explore applications in robotic systems, including sensory integration and rhythmic pattern generation for locomotion.

\section{Acknowledgments}
This work was funded by National Science Foundation (NSF) DBI 2015317 as part of the NSF/CIHR-/DFG/FRQ/UKRI-MRC Next Generation Networks for Neuroscience Program, as well as NSF FRR IIS-2138873. We would like to thank J. Stu McNeal and Clayton Jackson for feedback on an earlier version of this manuscript. 
ChatGPT was used to assist with minor programming tasks (e.g., utility functions), code formatting for figure preparation, and limited language editing for clarity. It was not used to generate any of the data, analyses, or findings presented in this work. The authors reviewed and verified all generated content, and take full responsibility for the content of the manuscript.

\section{Author Contributions}
Conceptualization, W.R.P.N. and R.D.Q.; methodology, W.R.P.N.; software, W.R.P.N.; validation, W.R.P.N.; formal analysis, W.R.P.N.; investigation, W.R.P.N.; resources, R.D.Q.; data curation, W.R.P.N.; writing—original draft preparation, W.R.P.N.; writing—review and editing, W.R.P.N. and R.D.Q.; visualization, W.R.P.N.; supervision, R.D.Q.; project administration, W.R.P.N. and R.D.Q.; funding acquisition, R.D.Q. All authors have read and agreed to the published version of the manuscript.

\section{Data Availability}
All code and data used to generate the figures will be made publicly available following publication.

\bibliographystyle{unsrt}
\bibliography{export.bib}

\end{document}